\documentclass[a4paper]{article}

\usepackage{INTERSPEECH2021}
\usepackage[implicit=false]{hyperref}

\usepackage{lipsum}
\usepackage{placeins}
\usepackage{adjustbox}
\usepackage{multirow}
\usepackage{tabularx}
\usepackage{amssymb}
    \newcolumntype{L}{>{\raggedright\arraybackslash}X}
    
\usepackage[dvipsnames]{xcolor}

\title{EasyCall corpus: a dysarthric speech dataset}
\name{Rosanna Turrisi$^{1,2,3}$, Arianna Braccia$^2$, Marco Emanuele$^{1,2}$, Simone Giulietti$^2$, Maura Pugliatti$^{2,4}$, \\Mariachiara Sensi$^{2,4}$, Luciano Fadiga$^{1,2*}$, Leonardo Badino$^{5*}$\\
$^*$These authors equally contributed to this work}
\address{
   $^1$Italian Institute of Technology,
  $^2$University of Ferrara,
  $^3$DIBRIS, University of Genoa,\\
  $^4$Neurology Unit, Azienda Ospedaliera Universitaria Sant'Anna, Ferrara,
  $^5$ PerVoice}
\email{trrrnn@unife.it, arianna.braccia@gmail.com, mnlmrc@unife.it, simone.giulietti5@gmail.com, maura.pugliatti@unife.it, mc.sensi@ospfe.it,  fdl@unife.it,  leonardo.badino@pervoice.it}

\begin{document}

\maketitle
\begin{abstract} 
This paper introduces a new dysarthric speech command dataset in Italian, called EasyCall corpus. The dataset consists of 
21386 audio recordings from 24 healthy and 31 dysarthric speakers, whose individual degree of speech impairment was
assessed by neurologists through the Therapy Outcome Measure. 
The corpus aims at providing a resource for the development 
of ASR-based assistive technologies for 
patients with dysarthria. In particular, it may be exploited to develop a voice-controlled contact application for commercial smartphones, aiming at improving dysarthric patients' ability to 
communicate with their family and caregivers. Before recording the dataset, participants were administered a survey to evaluate which commands are more likely to be employed
by dysarthric individuals in a voice-controlled contact application. In addition, the dataset includes a list of non-commands (i.e., words near/inside commands or phonetically close to commands) that can be leveraged to build a more robust command recognition system. 
At present commercial ASR systems perform poorly on the EasyCall Corpus as we report in this paper.
This result corroborates the need for dysarthric speech corpora for developing effective assistive technologies.
To the best of our knowledge, this database represents the richest corpus of dysarthric speech to date.
\end{abstract}

\section{Introduction}

In recent years, the outstanding improvement of Automatic Speech Recognition systems has been increasingly promoting users interaction with commercial electronic devices, such as smartphones. This may have important implications for patients with physical disabilities, possibly helping them to communicate
with their family and caregivers. As residual speech is often the last and most effective way of interacting with the external world, ASR systems tailored on these patients’ specific needs may substantially improve their quality of life. However, the recognition of utterances produced by individuals with speech impairments, such as dysarthria, may still be challenging for traditional ASR systems. \\ Dysarthria is a motor speech disorder quite common in elderly people and in conditions such as Parkinson Disease, Amyotrophic Lateral Sclerosis and post-stroke motor impairments. As the speech characteristics of a dysarthric speaker vary based on the type and 
severity of dysarthria, the ASR training requires a large amount of dysarthric speech data in order to capture 
such variability. However, only limited and small dysarthric speech corpora are 
currently available.\\

In this work, we introduce and describe a database of command speech 
recorded from healthy individuals and dysarthric patients. This corpus aims at providing a new resource for future developments of ASR-based assistive technologies. The recordings focus on a small vocabulary, including basic smartphone commands, such as “open contacts”, “start call”, “end call”. The reason that led us to focus on small vocabulary recordings is that, unfortunately, collecting large speech recordings from dysarthric speakers is often unfeasible for practical reasons (e.g. low compliance, fatigue during recordings). \\
On the other hand, ASR systems are typically based on neural networks that require a large amount of data and, roughly speaking, follow the rule that “more data imply better performance”. This is especially the case for large vocabulary tasks. For example, a previous work has shown that TED-LIUM 3 outperforms the model trained on TED-LIUM 2 dataset by doubling the amount of data \cite{hernandez2018ted}. Similarly, VoxCeleb 2 augmented the number of utterances from 100000 to one million, leading to an
increase of the ASR performance \cite{chung2018voxceleb2}. Focusing on small-vocabulary ASR reduces the model complexity and allows us to collect multiple examples of the same command. The drawback is that it forces us to choose a limited number of utterances that can be recognized. We therefore took into account the specific task of making a phone call, as this is the first and basic problem dysarthric patients may face to connect with other people through a smartphone. \\
The collected corpus is currently public and can be downloaded at \href{http://neurolab.unife.it/easycallcorpus/}{\textit{this link.}}\\

\section{Existing datasets}\label{sec:s1}

Only very few and limited corpora of Italian dysarthric speech have been collected to date. In \cite{ballati2018assessing}, the authors collected 34 sentences from 8 patients with Amyotrophic Lateral Sclerosis (ALS) to test the recognition performance of the most used virtual assistants (e.g., Apple’s Siri) on dysarthric speech.
A larger corpus is introduced in \cite{pettorino2016speech}. This contains read speech recorded from 15 neurologically healthy speakers and 11 individuals affected by Parkinson's Disease (PD). Nonetheless, all PD patients included in this study were diagnosed with mild dysarthria. Therefore, this dataset only accounts for a restricted range of dysarthria severity. 
Recently, speech recordings from 29 Italian native speakers have been collected within the AllSpeak project \cite{dinardi}. The dataset contains 25 commands in italian, relative to basic needs such as ``I am thirsty". Specifically, 2387 examples have been recorded from 17 individuals affected by Amyotrophic Lateral Sclerosis, and 1857 speech recordings are produced by  13 speakers of control. Unfortunately, this dataset is not 
public. 
\\

A British English dysarthric speech dataset, called The homeService corpus \cite{nicolao2016framework}, was gathered as part of the homeService project that aims at helping dysarthric individuals to interact with their home appliances through 
voice control. The majority of this corpus contains real user-device interactions, in which commands were freely chosen by the user and recorded in domestic environments. 
Despite this great advantage, the use of this corpus is limited as only 5 dysarthric patients were included.

Popular dysarthric speech corpora in American English are the TORGO dataset \cite{Rudzicz2012torgo}, the Nemours corpus \cite{menendez1996nemours}, and the Universal Access (UA) speech \cite{kim2008dysarthric}. 
The TORGO database consists of aligned acoustic and articulatory recordings from 15 speakers, including 7 control speakers without any disorder and 8 speakers presenting different levels of dysarthria. The subjects were asked to read single words or sentences and to describe the content of some photos. A total of 5980 and 2762 utterances were recorded from healthy and dysarthric speakers, resulting in approximately three hours of speech.  \\
The Nemours database is a collection  of 74 short sentences spoken by 11 speakers with varying degrees of dysarthria, resulting in a total number of 814 recordings. The sentences are nonsense phrases with a fixed structure, “the X is Y the Z”. 
Usually, Y is a verb in the present participle form, while X and Z are nouns. An example of sentence is “The shin is going the who”. \\
To the best of our knowledge, the UA-Speech database is the largest corpus of dysarthric speech in American English. It is a collection of 541 read speech recordings from 19 individuals with cerebral palsy. The prompt words include: three repetitions of the first ten digits, three repetitions of 26 radio alphabet letters, three repetitions of 19 computer commands, common words form the ``Grandfather Passage'' and uncommon words from phonetically balanced sentences (TIMIT \cite{zue1990}) one time each. \\

\section{Data collection}\label{sec:s2}
\subsection{Participants}
We recorded utterances from 31 dysarthric (11 females, 20 males) and 24 healthy (10 females, 14 males) speakers. The inclusion criteria for dysarthric speakers 
were:
\begin{itemize}
    \item age $\geq$ 18;
    \item dysarthria related to Parkinson's Disease, Huntingon's Disease, Amyotrophic Lateral Sclerosis,  peripheral neuropathy,  myopathic or myasthenic lesions.
\end{itemize}
Exclusion criteria were:
\begin{itemize}
    \item aphasic syndromes; 
    \item dementia;
    \item intellectual disability;
\end{itemize}
For each dysarthric speaker,
the type and the severity of the dysarthria were assessed by an experienced neurologist through the Therapy Outcome Measure (TOM) \cite{Enderby1997} (Table \ref{tab:easycall}). Specifically, the TOM score ranges from 1 to 5 corresponding to \textit{mild}, \textit{mild-moderate}, \textit{moderate}, \textit{moderate-sever}, and \textit{severe} dysarthria.\\

Participants provided their informed consent before undergoing the experimental procedures. The protocol was designed according to the declaration of Helsinki and approved by the local ethical committee.

\subsection{Procedures}
\begin{figure}[t]
\centering\includegraphics[width=0.99\linewidth]{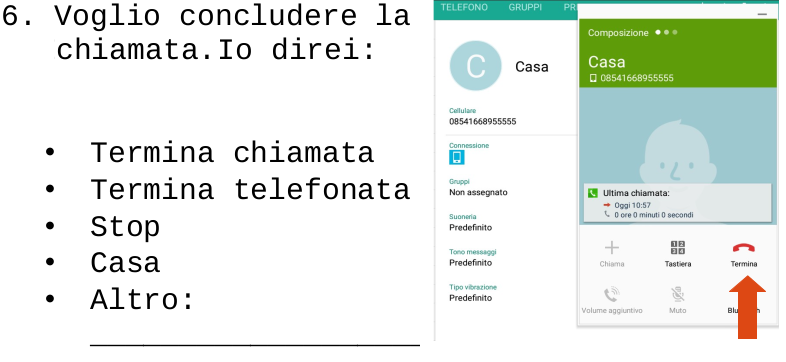}
\caption{An example question of the survey: the subject is asked which command she would use to conclude the call. 
Options are two Italian variants of ``end the call",
``stop",
a misleading command (``home") and
``other".}
\label{fig:survey}
\end{figure}

We collected speech recordings of commands related to the task ``make a call", 
aiming at creating a dataset that could be exploited, in the future, for building a smartphone application able to
\begin{itemize}
\item type and call phone numbers;
 \item save new contacts;
 
 \item provide additional options (e.g., rapid/favorite contacts, speakerphone, etc.).
\end{itemize}
The choice of the commands is of crucial importance to build a corpus suitable for developing assistive technologies in real-life contexts. 
Importantly, 
contacts 
often include 
a very large number of proper names, as well as common
contacts such as ``home", ``mum", ``dad". Recording all of them would be unfeasible. Thus, we reasoned that
a	hypothetical 
voice-controlled smartphone application 
could not allow to 
say, for instance, ``call Marco". Rather, the task may be accomplished through a step-by-step procedure. A reasonable sequence of commands could be ``open 
contacts", ``scroll down to the letter M", ``scroll down" (repeated until the desired contact is reached), ``call", ``end 
call".\\

Moreover, the design of a dysarthric speech database must take into account that speech-impaired patients may adopt uncommon words or expressions due to difficulties in pronouncing some words. In addition, the vocabulary of elderly persons (who are more likely to be affected by articulatory problems) may be different from young individuals due to cultural and generational influences. In order to record plausible commands, we first assessed 
which specific words dysarthric patients would most likely employ to accomplish the task of making a call. 
For this purpose, patients were asked to imagine themselves having to call the contact ``home" using an hypothetical voice-controlled smartphone,  without explicitly pronouncing the word ``home". This constraint was added to force patients 
to use a step-by-step procedure involving different types of commands (e.g. a sequence of commands similar to the one described above) 
to accomplish the task.   

\begin{table*}[t]
\caption{EasyCall Corpus. Controls and patients refer to healthy and dysarthric speakers, respectively. For both of them, we reported the number of recorded sessions and utterances. In addition, the TOM rating is reported for patients. In the speaker code, F/M stands for female/male, while the letter C is added to refer to control subjects. The mismatch between the number of wav files, with the same number of sessions, for some speakers depends on the fact that we updated the command list during the experiment time by adding new commands.}
\begin{center}
\begin{tabular}{c|c|c|c|c|c|c|c} 
\multicolumn{3}{c|}{\textbf{Controls}} & \multicolumn{5}{c}{\textbf{Patients}}\\
\hline
\textbf{Speaker code}& \textbf{N. Sessions} & \textbf{N. wav files} & \textbf{Speaker code} & \textbf{Type of dysarthria} & \textbf{TOM} & \textbf{N. Sessions} & \textbf{N. wav files} \\
\hline
FC01 & 6 & 396  & F01 & paretic & 1 & 5 & 330 \\
FC02 & 6 & 396 &  F02 & paretic & 3 & 6  & 396\\
FC03 & 6 & 396 &  F03 & paretic & 1 & 6  & 396 \\
FC04 &  6 & 396 & F04 & - & - & 4 & 234 \\
FC05 &  7 & 483 & F05 & paretic & 1 & 6 & 414\\
FC06 &  6 & 414 & F06 & paretic & 1 & 6 & 414\\
FC07 & 6  & 414 & F07 & paretic & 1 & 6 & 414\\
FC08 & 8 & 552  &  F08 & paretic & 1 & 6 & 414\\
FC09 & 6 & 414 & F09 & paretic & 1 & 6 & 414 \\
FC10 & 6 & 414 & F10 & paretic &  5 & 2 & 18\\
MC01 & 6 & 396& F11 & cerebellar & 2 & 5 & 345 \\
MC02 & 6  & 396 & M01 & extrapyramidal & 3 & 6 & 395 \\
MC03 & 6  & 396 & M02 & paretic & 1 & 6 & 396 \\
MC04 &  6  & 396&  M03 & paretic & 3 & 6  & 396\\
MC05 & 7  & 462 & M04 & paretic & 1 & 6 & 396 \\
MC06 & 6  & 396 & M05 & paretic & 4 & 6  & 396\\
MC07 & 7 & 462  & M06 & paretic & 4 & 6  & 396\\
MC08 & 6  & 414 & M07 & paretic & 4 & 6  & 396\\
MC09 & 6  & 414& M08 & paretic & 3 & 6  & 396\\
MC10 & 6  & 414& M09 & cerebellar & 1 & 6 & 396 \\
MC11 & 6  & 414 & M10 & paretic & 3 & 6 & 413\\
MC12 & 6  & 414  & M11 & paretic & 5 & 6 & 414\\
MC13 & 6  & 414  & M12 & cerebellar & 1 & 6 &414\\
MC14 & 6 & 414 & M13 & paretic & 1 & 6 & 414\\
\multicolumn{3}{c|}{} & M14 & pyramidal & 5 & 4 & 181\\
\multicolumn{3}{c|}{} & M15 & paretic & 1 & 6 & 414\\
\multicolumn{3}{c|}{} & M16 & paretic & 3 & 2 & 138\\
\multicolumn{3}{c|}{} & M17 & paretic & 1 & 6 & 414\\
\multicolumn{3}{c|}{} & M18 & paretic & 1 & 6 & 414\\
\multicolumn{3}{c|}{} & M19 & paretic & 3 & 3 & 207\\
\multicolumn{3}{c|}{} & M20 & cerebellar & 1 & 6 & 414\\
\end{tabular}
\end{center}
\label{tab:easycall}
\end{table*}
\FloatBarrier
Patients completed a survey including ten questions covering all the steps of the procedure and  some additional options (e.g., add a contact to the list of the favorite ones).
For each question, we suggested 3 plausible commands, 1 completely unrelated command and the option ``other", in which subjects provided their own answer.
In addition, a figure depicting the step related to each question was shown in the survey to promote a proper understanding of the task. An example question of the survey is shown in Fig. \ref{fig:survey}.

Audio recordings were obtained using a in-house  smartphone application. The application displayed the sentences that subjects were asked to read from a smartphone screen and simultaneously recorded the speech. 
To avoid patients' fatigue, 
recordings were split in sessions, during which one repetition of all commands was recorded. 
Audio recordings were automatically saved in the .wav format and kept anonymous. 
 Note that recording through a smartphone microphone allowed to closely represent the audio recorded in real-life contexts (e.g., in a voice command-based contact application), thus reducing the potential mismatch between experimental and actual conditions due to different recording tools.

\subsection{The dataset}
Based on the survey, we created and then updated the list of commands resulting in a final list of 67 sentences, including 
37 commands (i.e., words or sentences related to the task of interest) and 30 non-commands. A first version (v1) of the vocabulary did not include the command ``esci da rubrica" 
(i.e., an Italian variant of ``
exit contacts"), which was subsequently added to an update version (v2) . 

Non-commands
can be words near or inside commands (e.g., the non-command ``contacts'' is contained in the command ``start contacts'') or sentences phonetically close to commands (e.g., the non-command ``Tra'', that means ``between'', is close to the command ``Tre'', i.e. ``three''). These can be employed to build a more robust Voice Command Recognizer that better discerns between targets that sound similar.  

We recorded the speech commands from 31 dysarthric speakers and 24 healthy speakers, each of whom performed from 2 to 8 sessions. In a session, the speaker repeats once each command in the vocabulary (i.e., 66 recordings for v1 and 67 for v2). In addition, in v2, the command 
``start the contact application" has been chosen as key command and recorded three times per session, resulting in a final number of 69 recordings.
The total number of recording is 21386, of which 10077 are from healthy speakers and the remaining 11309 from dysarthric ones. \\
Table \ref{tab:easycall}  
details the data we collected so far. 

\vspace{1cm}

\begin{table}[t]
\caption{Evaluation of two commercial ASR systems on a subset of EasyCall corpus in terms of Word Error Rate.}
\begin{adjustbox}{width=.5\textwidth,center}
\begin{tabular}{ccc|cccc} 
\multicolumn{3}{c|}{\textbf{Controls}} & \multicolumn{4}{c}{\textbf{Patients}}\\
\hline
\textbf{Speaker code}& \textbf{ASR1} & \textbf{ASR2} & \textbf{Speaker code} & \textbf{TOM} & \textbf{ASR1} & \textbf{ASR2}\\
\hline
FC05 & 7.77 & 26.21 &F05 & 1  & 76.70& 157.28\\
FC06 & 3.88 & 11.65 &  F06& 1  & 42.72& 130.10\\
FC08 & 9.71 & 24.27 &  F07& 1 & 52.43& 106.80\\
FC09 & 3.33 & 28.89 & F08& 1 & 56.44& 158.42\\
FC010 & 1.98 &  6.93 &F09 & 1  & 60.40& 108.91\\
MC08 & 6.80& 26.21 & M10& 3  & 85.44& 100.97\\
MC09 & 4.90 & 15.69 & M11& 5 &91.18& 212.75\\
MC010 &9.80 &  47.06 &M12& 1 & 25.24& 87.38\\
MC011 & 0.98& 18.63 &M13& 1 &31.68& 78.22 \\
MC012 &4.85 & 34.95 & M14& 5  & 102.88& 254.81\\
MC013 & 17.48& 60.19 & M15&1 &34.62& 98.08\\
\multicolumn{3}{c|}{} &  M16 & 3& 82.52& 128.16\\
\hline
\textbf{Average} & 6.55 & 27.35 & \textbf{Average} & - & 61.90 & 135.26 \\
\end{tabular}
\end{adjustbox}
 \label{tab:asr}
\end{table}
\section{Experiments}\label{sec:s3}
We evaluated the collected dataset on some most known commercial ASR systems, that are Microsoft and IBM Speech-to-Text. 
Specifically, we selected a subset of healthy and dysarthric speakers with v2 vocabulary and we tested one repetition of each command.
Table \ref{tab:asr} reports the results in terms of Word Error Rate (WER). Even though the ASR systems achieve a low WER on healthy speech, they show a poor performance on dysarthric speech. Indeed, one of the two ASR systems (ASR1) achieves an average WER on dysarthric speech of $61.90\%$, while only $6.55\%$ on healthy speech, leading to an absolute difference of $55.35\%$. Similarly, the other system (ASR2) provides an average WER of $27.35\%$ and $135.26\%$ on healthy and dysarthric speakers, respectively, showing a mismatch between the two of  $107.91\%$. For both ASR systems, the misrecognition especially occurs for speakers M11 and M14, who have a severe dysarthria level, providing a WER always closer or higher than $100\%$. Note that a WER higher than $100\%$ also occurs in the ASR2 system for most of the dysarthric speakers. This is due do the speakers' mumbling that the ASR system attempts to recognize as a sequence of words. Another important factor that impacts the WER is the error of substitution, insertion or deletion done by the patients. 
Indeed, we did not use as true label the transcription of the speech as pronounced by the patient, but rather the target sequence of words (i.e., the command that the patient reads and attempts to state).  The reason underlying this choice is that ASR systems for impaired speech should be able to recognize what a patient wants to communicate rather than what she actually says, by managing the pronunciation errors intrinsic to the disease.  
For example, if patients have trouble pronouncing the combination of letters ``st", they could systematically 
mispronounce the word ``stop" by producing a sound similar to ``top". In this case, established ASR systems may correctly detect the phonetically most plausible word (i.e., ``top") and fail at tracing it back to the true label (i.e. ``stop").
These results indicate 
that standard ASR systems completely fail in presence of dysarthric speech and, consequently, the collected corpus represents a challenging dataset and a fundamental resource to develop ASR models robust to dysarthria.

\section{Conclusions}\label{sec:s4}
In this work, we introduce
the EasyCall corpus consisting of 
of 21386 speech commands, including 10077 recordings from 24 healthy subjects and 11309 from 31 dysarthric speakers. To the best of our knowledge, it represents the richest databases of speech recorded from subjects with dysarthria to date. 
The collection of data for this project is still ongoing and, accordingly,
this database is expected to expand 
in the future.\\

 We firmly believe that this corpus will provide a fundamental resource for developing assistive technologies for individuals with dysarthria. Indeed, this dataset can be exploited to train a voice command recognition system enabling dysarthric individuals to control a contacts mobile application through their voice and make calls in an easy, user-friendly way, tailored on their specific needs. This may result in a substantial increase in the communicative skills of speech-impaired patients, ultimately contributing to improve their quality of life. Moreover, the data recordings include non-commands phonetically close to commands (e.g., ``tre" is close to ``tra") or near/inside the commands (e.g., ``end" is included in ``end call"). The 
 inclusion of these audio tracks can improve the robustness of the grammar model. \\ 

Furthermore, 
the corpus described here could be exploited to pre-train a network for a different goal. Indeed, when the dataset size is limited, a common strategy to improve the ASR system performance is pre-training the network model 
using a larger dataset. Due to the lack of dysarthric speech corpora,  
this step is usually achieved by using healthy speech datasets. Pre-training a network with dysarthric speech data would bring a more representative model in which the training distribution is closer to the distribution of the data of interest.  


\bibliographystyle{IEEEtran}

\bibliography{mybib}


\end{document}